\pdfoutput=1

\documentclass[11pt]{article}

\usepackage[preprint]{acl}

\usepackage{times}
\usepackage{latexsym}

\usepackage[T1]{fontenc}

\usepackage[utf8]{inputenc}

\usepackage{microtype}

\usepackage{inconsolata}

\usepackage{graphicx}

\usepackage{multirow}
\usepackage{CJKutf8}
\usepackage[most]{tcolorbox}
\usepackage{booktabs}
\usepackage{bm}
\usepackage{makecell}

\definecolor{seagreen}{RGB}{71, 196, 160}
\definecolor{tealblue}{RGB}{17, 184, 184}
\definecolor{violetred}{RGB}{252, 90, 171}
\newcommand{\cyanbox}[1]{\colorbox{cyan!15}{#1}}
\newcommand{\tealbox}[1]{\colorbox{tealblue!20}{#1}}
\newcommand{\redbox}[1]{\colorbox{violetred!15}{#1}}

\title{Are Large Language Models Chronically Online Surfers?\\A Dataset for Chinese Internet Meme Explanation}

\author{Yubo Xie$^1$\footnotemark[1], Chenkai Wang$^2$, Zongyang Ma$^3$, Fahui Miao$^1$ \\
$^1$Shanghai Maritime University, Shanghai, China \\
$^2$École Polytechnique Fédérale de Lausanne, Lausanne, Switzerland \\
$^3$Xi'an Jiaotong Liverpool University, Suzhou, China \\
\texttt{yuboxie@hotmail.com, wangchenkaicn@foxmail.com,} \\
\texttt{Zongyang.Ma@xjtlu.edu.cn, miaofahui@126.com}}

\begin{document}
\maketitle
{
\renewcommand{\thefootnote}{\fnsymbol{footnote}}
\footnotetext[1]{Corresponding author.}
}
\begin{abstract}
Large language models (LLMs) are trained on vast amounts of text from the Internet, but do they truly understand the viral content that rapidly spreads online---commonly known as memes? In this paper, we introduce CHIME, a dataset for \textbf{CH}inese \textbf{I}nternet \textbf{M}eme \textbf{E}xplanation. The dataset comprises popular phrase-based memes from the Chinese Internet, annotated with detailed information on their meaning, origin, example sentences, types, etc. To evaluate whether LLMs understand these memes, we designed two tasks. In the first task, we assessed the models' ability to explain a given meme, identify its origin, and generate appropriate example sentences. The results show that while LLMs can explain the meanings of some memes, their performance declines significantly for culturally and linguistically nuanced meme types. Additionally, they consistently struggle to provide accurate origins for the memes. In the second task, we created a set of multiple-choice questions (MCQs) requiring LLMs to select the most appropriate meme to fill in a blank within a contextual sentence. While the evaluated models were able to provide correct answers, their performance remains noticeably below human levels. We have made CHIME public\footnote{\url{https://github.com/yuboxie/chime}} and hope it will facilitate future research on computational meme understanding.
\end{abstract}

\section{Introduction}
\begin{figure}[t]
    \centering
    \includegraphics[width=\linewidth]{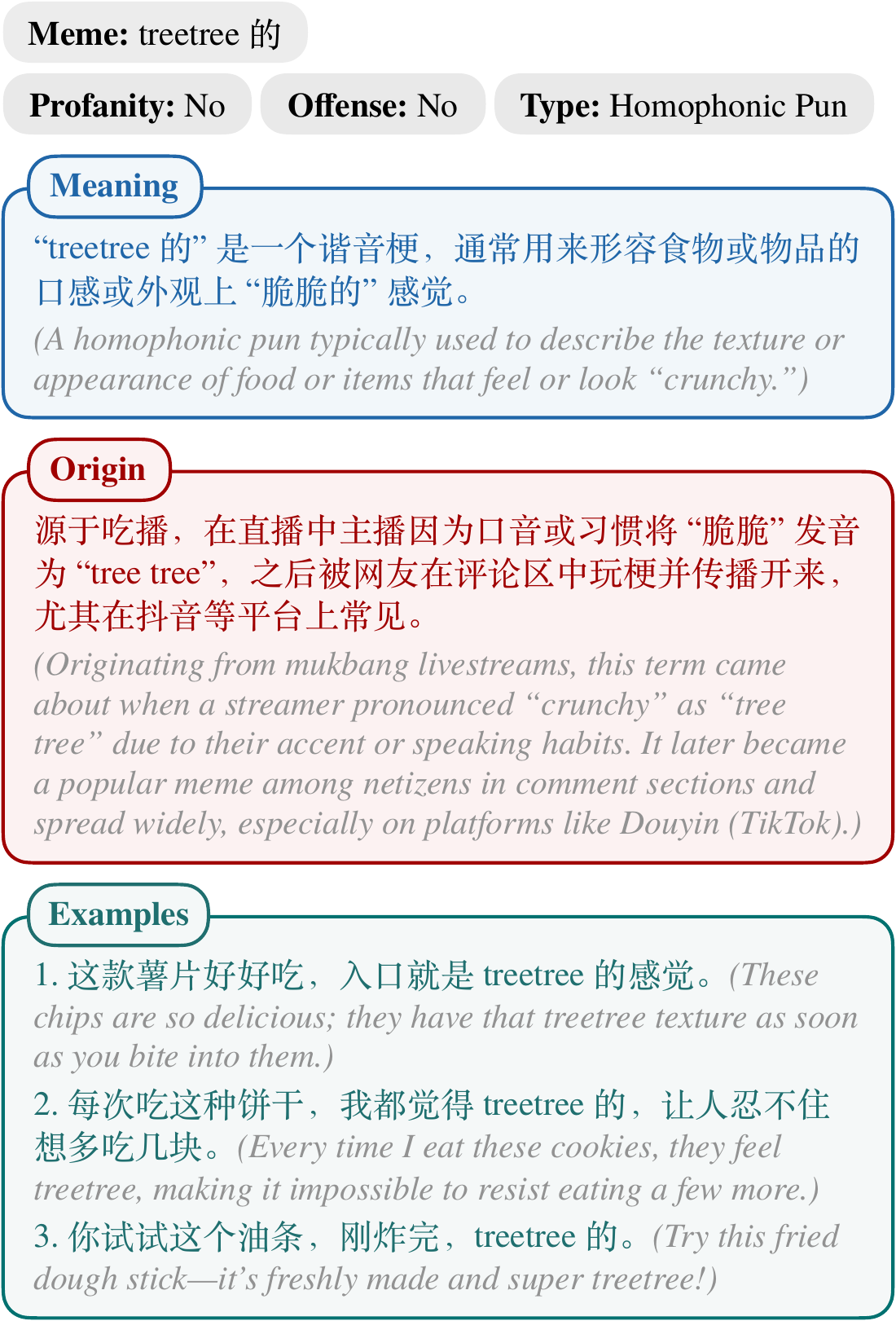}
    \caption{An example from our CHIME dataset.}
    \label{fig:chime_example}
\end{figure}
\begin{CJK*}{UTF8}{gbsn}
An Internet meme is a cultural item that conveys a specific idea, behavior, or style and spreads rapidly online, especially through social media and messaging platforms. While memes often gain popularity for their humorous and playful nature, they also reflect various facets of social, political, and cultural discourse~\citep{doi:10.1177/0920203X14531538,doi:10.1080/10447318.2022.2158260}. Internet memes take many forms, including phrases, images, and videos. In China, phrase-based memes have become a significant part of Internet culture, offering a distinctive blend of linguistic and cultural nuances. These phrases are typically short and straightforward. For example, some memes originate from slang (e.g., 熊孩子, ``\textit{brat}''), others are abbreviations (e.g., yyds/永远的神, ``\textit{the GOAT}'' or ``\textit{the greatest of all time}''), and some are created using phonetic transformations (e.g., 因缺思厅, ``\textit{interesting}'').
\end{CJK*}

Despite their playful appearance, Internet memes pose intriguing challenges for natural language understanding systems. They often rely on subtle wordplay, intertextual references, and constantly evolving cultural contexts, making them difficult even for humans to interpret without sufficient background knowledge~\cite{kostadinovska2018internet}. Specifically, Chinese Internet memes present unique challenges due to their use of puns, phonetic transformations, and extensive cultural references. Such memes frequently originate from online communities like Douyin (TikTok) and Weibo, where they can gain national attention in a matter of hours or days. Additionally, Chinese meme culture tends to blend homophones, dialect expressions, and creative abbreviations, resulting in content that is not only linguistically complex but also deeply rooted in shared social contexts. Recent advancements in large language models (LLMs)~\cite{gpt-4o,claude-3.5-sonnet,DBLP:journals/corr/abs-2407-21783,DBLP:journals/corr/abs-2406-12793,DBLP:journals/corr/abs-2412-15115,DBLP:journals/corr/abs-2412-19437} have shown promise in many natural language tasks, including conversational agents, information extraction, and machine translation. These models were pre-trained on vast amounts of text data from the Internet, which includes memes. However, whether these models can effectively capture the shifting and nuanced semantics of memes remains an open question.


To close this gap, we introduce the CHIME (\textbf{CH}inese \textbf{I}nternet \textbf{M}eme \textbf{E}xplanation) dataset---a collection of widely used simplified Chinese phrase-based memes, each annotated with detailed metadata on its meaning, origin, example usage, etc. (see Figure~\ref{fig:chime_example} for an example). Our goal is twofold. First, by assembling memes of varying linguistic complexity and cultural depth, CHIME serves as a resource to test whether LLMs can go beyond surface-level understanding. Second, by including annotations such as etymology and contextual usage, CHIME provides a more nuanced evaluation framework for computational meme comprehension. We posit that assessing how LLMs handle these memes offers fresh insights into the models' capabilities---and limitations---in reasoning about culturally rich, rapidly evolving content.

To this end, we propose two main tasks. The first task is an explanation-centric evaluation, where LLMs must describe a meme's meaning, provide its origin, and generate an appropriate example sentence. This setup probes both the breadth of the models' knowledge (e.g., recognizing the source and historical context of a meme) and the depth of their linguistic capabilities (e.g., producing example usage that aligns with social norms and cultural connotations). The second task is a multiple-choice question (MCQ) test, where the model must select the most fitting meme to fill in a blank within a contextual sentence. This requires not only semantic understanding but also the ability to discern subtle differences between multiple memes with overlapping or related meanings. Our findings suggest that while current LLMs can sometimes provide accurate meme explanations---especially for more straightforward or widely disseminated memes---their performance declines markedly for culturally and linguistically intricate cases. Furthermore, they struggle to pinpoint the correct origin of many memes, revealing gaps in their domain knowledge and context comprehension. By highlighting these challenges, we aim to spur further research in computational approaches for meme understanding, particularly those that incorporate cultural context into language models. We believe CHIME will pave the way for future investigations into how LLMs process and understand socially driven content on the Internet and contribute to the development of more humorous and human-like conversational agents.


\section{Related Work}

\subsection{Meme Datasets}
The concept of ``meme'' was first introduced by biologist Richard Dawkins in his book \textit{The Selfish Gene}~\cite{dawkins2016selfish}. The term ``Internet meme'' was formally defined by \citet{castano2013defining} as a phrase, image, or video associated with real-life events that spreads widely online.
Existing meme datasets mainly focus on image-based memes. \citet{DBLP:conf/nlpcc/LiLYXZ22} introduced a multimodal dataset for humor analysis using meme templates.
\citet{DBLP:conf/sigir/XuLZNZL022} introduced MET-Meme, a multimodal meme dataset rich in metaphorical features.
\citet{DBLP:conf/ijcnlp/HossainSH22, DBLP:conf/acl-trac/SuryawanshiCAB20} introduced multimodal meme datasets for identifying hateful and offensive content, while \citet{DBLP:conf/nips/LuXZWZZYL24,10720078} built multimodal Chinese harmful meme datasets. In our research, we develop a novel meme explanation dataset that focuses exclusively on text, with the goal of accurately explaining phrase-based memes.


\subsection{Non-Literal Language}
Non-literal language encompasses various forms of expression, including slangs, idioms, and figurative language. Several existing works have focused on the challenges of understanding non-literal language. \citet{DBLP:conf/acl/ZhengHS19,de-luca-fornaciari-etal-2024-hard} focus on idioms and their assessment in LLMs. \citet{DBLP:conf/naacl/LiuCZN22} assessed language models' ability to interpret figurative language by collecting creative metaphors from crowdsourcing workers. \citet{DBLP:conf/emnlp/MeiLWBC24} developed an English slang dataset from Urban Dictionary that reflects Internet language trends. Our dataset differs not only in language but also in how memes and slangs are created: Chinese Internet memes frequently utilize phonetic wordplay and visual puns, whereas English slangs typically rely on Latin letters and tend to favor abbreviations and acronyms. \citet{DBLP:conf/naacl/SunH0Z024} also constructed an English slang dataset, but primarily from movie subtitles, which may not capture the most recent Internet language trends.

Some other works have focused on toxic and offensive language detection, which may contain Internet slangs. \citet{DBLP:conf/acl/LuXZMYL23} constructed a fine-grained dataset and insult lexicon to detect Chinese toxic language. \citet{DBLP:conf/emnlp/XiaoHCL24} evaluated the robustness of language models in detecting disguised Chinese offensive content.

\subsection{Humor Datasets}
Humor is defined as the tendency of experiences to evoke laughter and provide amusement. Traditionally, humorous content has been represented as plain text. \citet{DBLP:conf/cikm/ZhangL14} developed a humor recognition model to identify humorous tweets.
\citet{DBLP:conf/emnlp/YangLDH15,DBLP:conf/emnlp/WellerS19,DBLP:conf/lrec/WellerS20} introduced various English humor datasets. \citet{DBLP:journals/corr/abs-2412-17729} introduced Chumor, a Chinese humor dataset sourced from Ruo Zhi Ba.
\citet{DBLP:conf/aaai/ChenYLLGGPLLX24} proposed TalkFunny, a Chinese explainable humorous response dataset.
Recent studies have also focused on multimodal humor datasets. \citet{DBLP:conf/emnlp/HasanRZZTMH19,DBLP:conf/nlpcc/WuLYX21,DBLP:conf/lrec/RadevSTPICJVJJM16,DBLP:conf/acl/HesselMHLDZM023} constructed and analyzed humor datasets from various sources like TED videos, TV sitcoms, and The New Yorker cartoons.
Our research focuses on Chinese phrase-based memes, which are a unique form of humorous content and have been rarely explored in existing literature.

\section{Dataset}
The CHIME dataset was developed by collecting human-written meme explanations from online sources, followed by the automatic extraction of key information and subsequent manual verification. Each entry in the dataset is manually annotated with labels for meme type and the presence of profanity and offensive content. The following subsections provide a detailed explanation of these processes.

\subsection{Raw Data Collection}
\begin{CJK*}{UTF8}{gbsn}
We first collected human-written meme explanations from Geng Baike (梗百科, \textit{Meme Encyclopedia})\footnote{\url{https://gengbaike.cn/}}, a website where users can contribute articles explaining specific phrase-based memes popular on the Chinese Internet. The explanations collected were created between August 17, 2020, and September 23, 2024. The data were then cleaned by correcting typographical errors and removing duplicates.
\end{CJK*}

To filter out memes that are too niche, five annotators (three of the authors and two recruited individuals) reviewed all the collected meme explanations, indicating whether they were familiar with each one. The annotators, all frequent Internet users with adequate digital literacy, represent a range of birth years from the 1980s to the 2000s. We retained only those memes recognized by at least one of the five annotators. This process resulted in a final collection of 1,458 meme explanations.

\subsection{Key Information Extraction}
Since the crawled meme explanations were written by different individuals, they vary in format and style. To ensure consistency and extract relevant information, we utilized a large language model (LLM) to automatically identify and extract key elements from the explanations. Specifically, we focused on the following aspects:
\begin{itemize}
    \item \textbf{Meaning}: A concise explanation of the meme, provided in a few sentences.
    \item \textbf{Origin}: The source of the meme, such as a famous movie, a celebrity quote, a TV show, or other cultural references. This information is included when available but is optional.
    \item \textbf{Examples}: For each meme, we extract up to three example sentences illustrating its usage. If the original explanation does not include examples, the LLM generates them.
\end{itemize}
We asked GPT-4o~\cite{gpt-4o} to extract the three components described above from each crawled meme explanation, using the prompt in Appendix~\ref{subsec:extraction_prompt}. However, the output of GPT-4o was not always fully accurate or reliable, as LLMs are known to generate erroneous or unfaithful content, commonly referred to as hallucinations~\citep{DBLP:journals/corr/abs-2311-05232}. Additionally, some of the extracted examples were generated by GPT-4o rather than originating from human-written explanations. As a result, we manually reviewed all extracted information to ensure the accuracy of the meanings and origins, verify that no key details were omitted, and confirm that the examples appropriately demonstrated the usage of each meme.

\subsection{Manual Annotation}
To ensure the dataset meets safety and ethical standards, each meme was manually annotated with two labels: a \textbf{profanity} label, indicating the presence of sexually explicit content, and an \textbf{offense} label, marking content that may be offensive, such as racism or discrimination. One of the authors conducted the initial annotation, which was then verified by the other two authors. Additionally, each meme was classified into one of the following types, based on a predefined taxonomy:
\begin{CJK*}{UTF8}{gbsn}
\begin{itemize}
    \item \textbf{Experience} (现象): Memes derived from individuals summarizing their personal experiences or situations. These are often used to express limitations or unmet expectations, serving as a form of self-relief or self-deprecation.
    \item \textbf{Quotation} (引用): Memes originating from historical stories, public events, movie plots, TV shows, or celebrity quotes.
    \item \textbf{Stylistic device} (修辞): Memes crafted using rhetorical techniques such as metaphor, irony, or sarcasm, often to convey auxiliary ideas or emotions.
    \item \textbf{Homophonic pun} (谐音): Memes created by replacing original characters with those of similar or identical sounds to produce humorous or meaningful effects.
    \item \textbf{Slang} (俗语): Memes based on widely recognized and popular colloquial expressions specific to a particular time or place.
    \item \textbf{Abbreviation} (缩写): Memes formed by shortening proper nouns or general phrases. The abbreviation methods vary and include morpheme reductions, initialisms, and simplified spellings.
\end{itemize}
\end{CJK*}
More details on the manual annotation process can be found in Appendix~\ref{subsec:manual_annotation}.

Table~\ref{tab:chime_dataset} presents the statistical overview of the CHIME dataset. Appendix~\ref{subsec:origin_statistics} provides additional statistics on the origins of the memes. We also provide a few representative examples for all six meme types in Appendix~\ref{subsec:chime_examples}.
\begin{table}[t]
    \centering
    \begin{tabular}{ll}
        \toprule
        \# Profanity & 75 (5.1\%) \\
        \# Offense & 127 (8.7\%) \\
        \midrule
        \# Experience & 561 (38.5\%) \\
        \# Quotation & 438 (30.0\%) \\
        \# Stylistic device & 214 (14.7\%) \\
        \# Homophonic pun & 133 (9.1\%) \\
        \# Slang & 60 (4.1\%) \\
        \# Abbreviation & 52 (3.6\%) \\
        \midrule
        \# Total & 1,458 \\
        \bottomrule
    \end{tabular}
    \caption{Statistical overview of the CHIME dataset.}
    \label{tab:chime_dataset}
\end{table}

\section{Can LLMs Explain Memes?}
The CHIME dataset functions as a benchmark for evaluating LLMs' capacity to interpret and explain memes without fine-tuning. To investigate this capability, we tasked candidate models with generating explanations for memes from this dataset.

\subsection{Experimental Setup}
We employ a zero-shot setting, prompting the candidate language models to explain the meaning of a given Internet meme, provide its origin (if available), and construct an example sentence. The prompts used can be found in Appendix~\ref{subsec:zero_shot_prompts}. We also experimented with one-shot prompting, but the results were mostly inferior to zero-shot prompting (see Appendix~\ref{subsec:one_shot_prompts}). The evaluated language models include GPT-4o~\cite{gpt-4o}, Claude 3.5 Sonnet~\cite{claude-3.5-sonnet}, GLM-4-9B, GLM-4-Plus~\cite{DBLP:journals/corr/abs-2406-12793}, Qwen2.5-7B, Qwen2.5-72B~\cite{DBLP:journals/corr/abs-2407-10671,DBLP:journals/corr/abs-2412-15115}, and DeepSeek-V3~\cite{DBLP:journals/corr/abs-2412-19437}.


\begin{table*}[t]
    \centering
    \begin{tabular}{lccccccccc}
        \toprule
        && \multicolumn{2}{c}{\textbf{Cosine Similarity}} && \multicolumn{2}{c}{\textbf{BERTScore (F)}} && \multicolumn{2}{c}{\textbf{BARTScore (F)}} \\
        \cmidrule{3-4} \cmidrule{6-7} \cmidrule{9-10}
        \textbf{Model} && \textbf{Meaning} & \textbf{Origin} && \textbf{Meaning} & \textbf{Origin} && \textbf{Meaning} & \textbf{Origin} \\
        \midrule
        GPT-4o && 0.805 & 0.628 && 0.790 & 0.680 && $-$4.367 & $-$4.695 \\
        Claude 3.5 Sonnet && 0.773 & 0.614 && 0.776 & 0.679 && $-$4.559 & $-$4.877 \\
        GLM-4-9B && 0.792 & 0.640 && 0.785 & 0.696 && $-$4.321 & $-$4.493 \\
        GLM-4-Plus && \cyanbox{\textbf{0.832}} & 0.689 && \cyanbox{\textbf{0.809}} & \cyanbox{\textbf{0.744}} && \cyanbox{$\bm{-}$\textbf{4.238}} & \cyanbox{$\bm{-}$\textbf{4.423}} \\
        Qwen2.5-7B && 0.778 & 0.579 && 0.765 & 0.632 && $-$4.448 & $-$4.855 \\
        Qwen2.5-72B && 0.805 & 0.622 && 0.789 & 0.676 && $-$4.321 & $-$4.602 \\
        DeepSeek-V3 && 0.787 & \cyanbox{\textbf{0.694}} && 0.782 & 0.740 && $-$4.289 & $-$4.463 \\
        \bottomrule
    \end{tabular}
    \caption{Average cosine similarity, BERTScore, and BARTScore across all six meme types for each candidate model. The best-performing scores are highlighted in \cyanbox{\textbf{bold}}.}
    \label{tab:auto_eval_avg}
\end{table*}
\begin{figure*}
    \centering
    \includegraphics{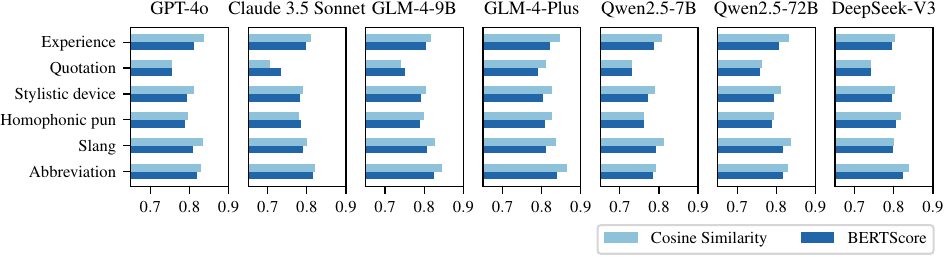}
    \caption{Average cosine similarity and BERTScore for the generated meanings of the candidate models, evaluated across each of the six meme types.}
    \label{fig:auto_eval_type}
\end{figure*}

\begin{table*}[t]
    \centering
    \begin{tabular}{lcccccccccccc}
        \toprule
        && \multicolumn{3}{c}{\textbf{Meaning (\%)}} && \multicolumn{3}{c}{\textbf{Origin (\%)}} && \multicolumn{3}{c}{\textbf{Example (\%)}} \\
        \cmidrule{3-5} \cmidrule{7-9} \cmidrule{11-13}
        \textbf{Model} && \textbf{A} & \textbf{N} & \textbf{D} && \textbf{A} & \textbf{N} & \textbf{D} && \textbf{A} & \textbf{N} & \textbf{D} \\
        \midrule
        GPT-4o && 53.9 & 9.0 & 37.1 && 18.5 & 8.2 & 73.3 && 55.0 & 8.3 & 36.7 \\
        Claude 3.5 Sonnet && 51.0 & 9.7 & 39.3 && 14.4 & 10.2 & 75.4 && 51.7 & 7.5 & 40.8 \\
        GLM-4-9B && 40.4 & 9.0 & 50.6 && 7.7 & 10.3 & 82.0 && 41.1 & 6.0 & 52.9 \\
        GLM-4-Plus && 68.5 & 8.9 & 22.6 && \tealbox{\textbf{35.9}} & 8.7 & 55.4 && 70.7 & 5.6 & 23.7 \\
        Qwen2.5-7B && 33.9 & 11.4 & 54.7 && 9.7 & 6.2 & 84.1 && 34.0 & 9.9 & 56.1 \\
        Qwen2.5-72B && 45.7 & 10.0 & 44.3 && 14.4 & 10.2 & 75.4 && 46.8 & 6.8 & 46.4 \\
        DeepSeek-V3 && \tealbox{\textbf{73.6}} & 10.3 & \tealbox{\textbf{16.1}} && 35.4 & 12.3 & \tealbox{\textbf{52.3}} && \tealbox{\textbf{77.4}} & 6.2 & \tealbox{\textbf{16.4}} \\
        \bottomrule
    \end{tabular}
    \caption{Average percentage of human ratings assigned as \textit{Agree}, \textit{Neutral}, and \textit{Disagree} across all six meme types for each candidate model. A stands for \textit{Agree}, N stands for \textit{Neutral}, and D stands for \textit{Disagree}. The best-performing scores are highlighted in \tealbox{\textbf{bold}}.}
    \label{tab:human_eval_results}
\end{table*}
\begin{figure*}
    \centering
    \includegraphics{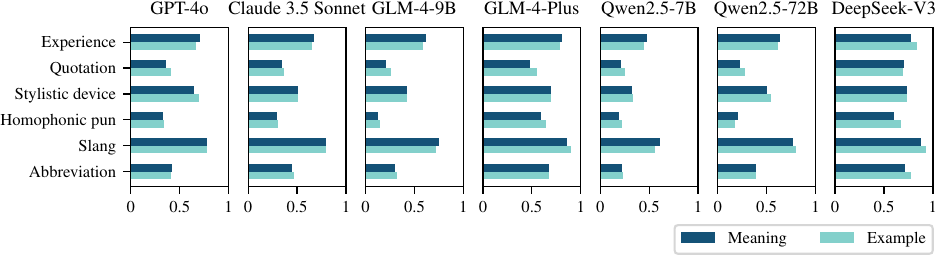}
    \caption{Average percentage of human ratings assigned as \textit{Agree} for the generated meanings and example sentences of the candidate models, evaluated across each of the six meme types. The results of the origin task are omitted, as most memes with an identifiable origin belong to the \textit{quotation} type.}
    \label{fig:human_eval_type}
\end{figure*}

\subsection{Automatic Evaluation}
Automatic evaluation was conducted on the entire dataset (1,458 memes), wherein LLM-generated interpretations of meme meaning and origin were systematically compared against the ground truth. We adopted the following metrics: cosine similarity, BERTScore~\cite{DBLP:conf/iclr/ZhangKWWA20}, and BARTScore~\cite{DBLP:conf/nips/YuanNL21}. For cosine similarity and BERTScore, we used the BGE embedding model (\textit{bge-large-zh-v1.5})~\cite{DBLP:conf/sigir/XiaoLZMLN24} to generate embeddings. For BARTScore, we used \textit{bart-large-chinese}~\cite{DBLP:journals/chinaf/ShaoGLDYYZBQ24}.

\paragraph{Overall Results}
Table~\ref{tab:auto_eval_avg} presents the average cosine similarity, BERTScore, and BARTScore across all six meme types for each of the six candidate models.\footnote{Since the BGE model was fine-tuned using contrastive learning, the absolute values of cosine similarity and BERTScore may not directly reflect performance quality; instead, the relative rankings are more informative.} As shown in the table, GLM-4-Plus achieves the highest scores on most metrics, while DeepSeek-V3 achieves the highest score on the origin task with cosine similarity. Additionally, all models perform better on the meaning task compared to the origin task, suggesting that identifying a meme's origin is more challenging than explaining its meaning. When comparing models of different sizes within the same series (e.g., GLM-4-9B versus GLM-4-Plus and Qwen 2.5-7B versus Qwen 2.5-72B), we observed that larger models consistently outperform their smaller counterparts.

\paragraph{Meme Type Specific Results}
Figure~\ref{fig:auto_eval_type} provides a detailed breakdown of meaning scores (cosine similarity and BERTScore) for each of the six meme types. Among these types, \textit{quotation} and \textit{homophonic pun} emerge as the most challenging to explain. For exact meaning scores for each meme type, refer to Appendix~\ref{subsec:auto_eval_results}.

\subsection{Human Evaluation}
To provide a more comprehensive and accurate assessment of the candidate models' performance---particularly for the generated example sentences, which cannot be effectively evaluated through automated methods---we conducted a human evaluation. We recruited individuals to rate the content generated by the language models. For each testing meme, raters were first shown the true meaning, origin (if available), and three example sentences. Then, for each of the seven candidate models, raters were asked to evaluate the generated meaning, origin (if available), and example sentences using a 3-point Likert scale based on the following statements:
\begin{enumerate}
    \item The explanation is completely accurate and aligns perfectly with the actual \textbf{meaning} of the meme. (\textit{Disagree}, \textit{Neutral}, \textit{Agree})
    \item The provided \textbf{origin} perfectly matches the source of the meme without any discrepancies. (\textit{Disagree}, \textit{Neutral}, \textit{Agree})
    \item The \textbf{example sentence} accurately reflects the actual usage of the meme, clearly and effectively demonstrating its meaning. (\textit{Disagree}, \textit{Neutral}, \textit{Agree})
\end{enumerate}
We randomly selected 240 testing memes (40 per category) and then divided them into 12 batches, each containing 20 memes for evaluation. For each batch, ratings were collected from three independent raters. More details on the human evaluation process are provided in Appendix~\ref{subsec:human_eval_instruct_example}.

\paragraph{Overall Results}
For each group of meme evaluation tasks, we calculated the Fleiss' kappa score to assess inter-annotator agreement. The average Fleiss' kappa score across all 12 groups is 0.442, indicating moderate agreement among the raters. The results of the human evaluation are presented in Table~\ref{tab:human_eval_results}, which shows the average percentage of ratings assigned as \textit{Agree}, \textit{Neutral}, and \textit{Disagree} for each model, based on the aspects of meaning, origin, and example sentence. Different from the automatic evaluation results, DeepSeek-V3 demonstrates the best performance on the meaning and example tasks. All models perform significantly worse on the origin task compared to the meaning and example tasks, and larger models generally outperform their smaller counterparts.

\paragraph{Meme Type Specific Results}
Figure~\ref{fig:human_eval_type} provides a comparison of all models' performance across the six meme types, showing the percentage of \textit{Agree} ratings for the meaning and example tasks. A strong correlation is observed between these two tasks, indicating that a model capable of accurately explaining the meaning of a meme is also likely to generate appropriate example sentences. Similar to the automatic evaluation results, \textit{quotation}, \textit{homophonic pun}, and \textit{abbreviation} are identified as the most challenging meme types to explain.

Additional details on the human evaluation results are provided in Appendix~\ref{subsec:human_eval_results}.

\subsection{Discussion}
Both automatic and human evaluations reveal significant variation in the performance of LLMs across different types of memes. While the models perform relatively well on \textit{experience} and \textit{slang} memes, their performance on \textit{quotation}, \textit{homophonic pun}, and \textit{abbreviation} memes is considerably lower. This disparity likely stems from the nature of these meme types: \textit{experience} memes often convey their meanings more directly, and \textit{slang} memes are typically well-known expressions used in local dialects, making them more prevalent in training data. In contrast, understanding \textit{quotation} memes often requires knowledge of their origin and contextual usage, while \textit{homophonic pun} and \textit{abbreviation} memes involve complex linguistic features that are harder to interpret at first glance. These findings suggest that comprehending memes with strong cultural and linguistic nuances remains a challenging task for LLMs, despite their advancements in overall language processing.

Though both evaluation methods indicate that GLM-4-Plus and DeepSeek-V3 are the two best-performing models, the rankings of the remaining models differ between automatic and human evaluations. Additionally, automatic metrics provide limited discriminatory power, as the scores among models are often quite close. While these metrics offer a quantitative measure of performance, they fail to capture subtleties such as contextual consistency and appropriateness in the generated content. The human evaluation results underscore the importance of incorporating qualitative assessments, particularly for tasks that demand nuanced understanding.

\paragraph{Error Analysis}
To further investigate the performance of LLMs, we conducted an error analysis on the generated meanings and origins. We have identified several consistent patterns: (1)~\textbf{Origin confusion}: Models frequently attributed memes to incorrect sources, particularly with \textit{quotation} memes. In many instances, LLMs provided vague attributions (e.g., ``originating from social media'') rather than specific origins. (2)~\textbf{Semantic shift}: For most misinterpreted \textit{homophonic pun} memes, models explained related concepts with similar phonetics rather than capturing the actual meme meaning. In other cases, models failed to recognize the phonetic wordplay entirely and simply explained the literal meaning. (3)~\textbf{Cross-type confusion}: \textit{Abbreviation} memes were occasionally misinterpreted as homophonic puns, indicating difficulty in distinguishing between these distinct linguistic mechanisms. We provide a more comprehensive error analysis with illustrative case studies in Appendix~\ref{subsec:error_analysis}.

\section{Can LLMs Use Memes?}
To evaluate LLMs' comprehensive meme literacy, we designed a second experiment where models must select the most appropriate meme to complete a contextual sentence with an intentional omission.

\begin{table*}[t]
    \centering
    \begin{tabular}{lcccccc|c}
        \toprule
        \textbf{Model} & {\small\textbf{Experience}} & {\small\textbf{Quotation}} & \makecell[cc]{\small\textbf{Stylistic}\\[-0.25em]\small\textbf{Device}} & \makecell[cc]{\small\textbf{Homophonic}\\[-0.25em]\small\textbf{Pun}} & {\small\textbf{Slang}} & {\small\textbf{Abbreviation}} & {\small\textbf{Average}} \\
        \midrule
        GPT-4o & 0.779 & 0.708 & 0.761 & 0.549 & 0.858 & 0.750 & 0.734 \\
        Claude 3.5 Sonnet & 0.758 & 0.644 & 0.778 & 0.597 & 0.800 & 0.729 & 0.718 \\
        GLM-4-9B & 0.574 & 0.527 & 0.536 & 0.360 & 0.654 & 0.504 & 0.526 \\
        GLM-4-Plus & 0.784 & 0.748 & 0.817 & 0.640 & 0.804 & 0.792 & 0.764 \\
        Qwen2.5-7B & 0.602 & 0.520 & 0.524 & 0.294 & 0.642 & 0.512 & 0.516 \\
        Qwen2.5-72B & 0.733 & 0.691 & 0.691 & 0.486 & \redbox{\textbf{0.869}} & 0.671 & 0.690 \\
        DeepSeek-V3 & \redbox{\textbf{0.831}} & \redbox{\textbf{0.791}} & \redbox{\textbf{0.828}} & \redbox{\textbf{0.713}} & 0.858 & \redbox{\textbf{0.833}} & \redbox{\textbf{0.809}} \\
        \midrule
        Human (Average) & 0.933 & 0.825 & 0.833 & 0.883 & 0.950 & 0.892 & 0.886 \\
        Human (Best) & 0.950 & 0.850 & 0.925 & 0.900 & 0.950 & 0.900 & 0.913 \\
        \bottomrule
    \end{tabular}
    \caption{Accuracy of the candidate models on the multiple-choice questions, along with human performance. The best-performing scores of the models are highlighted in \redbox{\textbf{bold}}.}
    \label{tab:accuracy}
\end{table*}
\begin{table}[t]
    \centering
    \begin{tabular}{lcc}
        \toprule
        \textbf{Model} & \textbf{Accuracy} & \bm{$\Delta$} \\
        \midrule
        GPT-4o & 0.896 & +0.162\\
        Claude 3.5 Sonnet & 0.881 & +0.163 \\
        GLM-4-9B & 0.692 & +0.166 \\
        GLM-4-Plus & 0.887 & +0.123 \\
        Qwen2.5-7B & 0.786 & +0.270 \\
        Qwen2.5-72B & 0.881 & +0.191 \\
        DeepSeek-V3 & \redbox{\textbf{0.897}} & +0.088 \\
        \bottomrule
    \end{tabular}
    \caption{Accuracy of the candidate models on the multiple-choice questions, \textbf{where the meaning of each meme option was provided to the LLMs}. The best-performing score is highlighted in \redbox{\textbf{bold}}. The column $\Delta$ indicates the improvement in accuracy compared to the setting without meme meanings (Table~\ref{tab:accuracy}).}
    \label{tab:accuracy_with_meaning}
\end{table}

\subsection{Experimental Setup}
In this experiment, we created a set of multiple-choice questions (MCQs) to evaluate the ability of candidate LLMs to select the most appropriate meme to complete a blank in a contextual sentence. Specifically, for each meme in the CHIME dataset, we randomly selected one of its example sentences and masked the targeted meme. We then identified four other memes with the highest cosine similarity, based on BGE embeddings, to serve as distractor options in the MCQ. As a result, the final testing set contains 1,268 MCQs.\footnote{The number of MCQs is less than the total number of memes because we used strict matching for masking targeted memes, but certain example sentences employ memes in a contextually flexible way.}

For each MCQ, the candidate models were prompted to choose the most appropriate meme from the given options. The prompt used is provided in Appendix~\ref{subsec:mcq_prompts_without_meaning}. Each MCQ was presented to the models five times, with the final prediction determined by majority voting. To mitigate potential biases in LLMs toward specific answer positions~\cite{DBLP:conf/iclr/Zheng0M0H24,DBLP:conf/acl/Sabour0ZLZSLMH24}, we further shuffled the order of the answer choices in four additional permutations, repeating the prediction process for each permutation. The average accuracy across these five runs was reported.

\subsection{Results}
Table~\ref{tab:accuracy} presents the accuracy of the candidate models on the MCQs, along with human performance. The results show that DeepSeek-V3 achieves the highest accuracy among the candidate models, outperforming the other models across all six meme types except \textit{slang}. The accuracy of the models varies significantly across different meme types, with \textit{experience} and \textit{slang} memes yielding higher accuracy compared to \textit{stylistic device} and \textit{homophonic pun} memes. As expected, larger models generally perform better than smaller models. The human performance, obtained from three recruited individuals on 240 randomly selected MCQs (balanced across meme types), serves as a general upper bound, with the average accuracy of human raters surpassing that of the models. The best human performance is also provided for reference.

\subsection{Discussion}
The results of the MCQ experiment demonstrate that LLMs can effectively leverage their learned knowledge to select the most appropriate meme to complete a contextual sentence. However, the accuracy of the models varies across different meme types, with models performing much worse on linguistically more nuanced memes such as \textit{homophonic pun}. This discrepancy is consistent with the findings from the meme explanation task, suggesting that the complexity of meme types significantly impacts the interpretive capabilities of LLMs.

We also conducted an experiment where the meaning of each meme option was provided to the LLMs, aiming to evaluate the impact of additional context on the models' performance (prompt provided in Appendix~\ref{subsec:mcq_prompts_with_meaning}). Table~\ref{tab:accuracy_with_meaning} presents the results in this setting. When the meaning of each meme option was provided to the models, the accuracy of all models increased, with the gap between the models narrowing. This finding suggests that LLMs can benefit from additional context to enhance their understanding and selection of memes, particularly for memes that involve complex linguistic features or cultural references.

To further understand the relationship between explanation and usage capabilities, we conducted cross-task analysis and found interesting patterns: (1) Models that correctly explain meme meanings achieve around 83\% accuracy in MCQ selection; (2) Conversely, models that select correct memes in context only achieve around 73\% accuracy in explanation. This asymmetry suggests that receptive understanding (recognizing appropriate usage) is easier than productive understanding (generating explanations), highlighting the distinct cognitive demands of these two tasks.

\section{Conclusion}
This paper introduces CHIME, a novel dataset designed for the explanation of Chinese Internet memes. Each meme in the dataset is annotated with detailed information, including its meaning, origin, example sentences, and auxiliary labels, creating a robust benchmark for evaluating and enhancing the interpretive capabilities of LLMs. Through a comprehensive experimental framework, we evaluated the performance of seven prominent LLMs, uncovering significant variability in their ability to explain memes across different types. In addition, we designed a multiple-choice question (MCQ) experiment in which models select the most appropriate meme to complete a contextual sentence, further highlighting the challenges in computational meme understanding, particularly for culturally and linguistically nuanced content. Future work could explore expanding the dataset to include multimodal memes and developing models that deliver more engaging and human-like conversational experiences with the support of the CHIME dataset.

\section{Limitations}
While the CHIME dataset provides a comprehensive benchmark for evaluating the interpretive capabilities of LLMs, it has several limitations. First, the dataset is limited to Chinese Internet memes, which may not fully represent the diversity of memes across different cultures and languages. Particularly, our dataset focuses on Simplified Chinese, because the source platform (Geng Baike) primarily hosts Simplified Chinese content and most Chinese Internet memes originate from mainland China platforms (Douyin, Weibo) where Simplified Chinese dominates. Future work could explore Traditional Chinese memes from Taiwan/Hong Kong platforms. Second, the dataset focuses on textual content, excluding multimodal memes that incorporate images, videos, or other media. Third, the reliance on human annotations introduces potential subjectivity and bias, and the limited number of annotators may affect the consistency of labeling. Lastly, the dataset captures memes from a specific time period, so its relevance may diminish as meme culture rapidly evolves. Future work could address these limitations by expanding the dataset to include a broader range of meme types and modalities, increasing annotation diversity, and continually updating the dataset to reflect the dynamic nature of meme culture.

\section{Ethical Considerations}
The CHIME dataset was created with the utmost care to ensure that all content is safe and appropriate for research purposes. We conducted manual annotation to identify and label any potentially offensive or inappropriate content, including profanity and discriminatory language. We acknowledge that Internet memes can sometimes perpetuate harmful stereotypes or biases, and we have taken care to document these occurrences through our labeling system to enable responsible research. We also considered the privacy implications of including user-generated content and took steps to anonymize any personally identifiable information.

The broader impacts of this work are both positive and potentially concerning. On the positive side, this dataset can help advance our understanding of how cultural information spreads online and how language models process culturally-embedded content. It may also aid in developing more culturally aware AI systems. However, we acknowledge potential risks, such as the dataset being used to generate misleading content or manipulate online discourse. We encourage researchers using our dataset to consider these ethical implications and implement appropriate safeguards in their work.


\bibliography{references}

\appendix

\section{Computing Infrastructure}
\label{sec:computing_infra}
All the experiments were conducted by invoking the models through their official APIs, with default hyperparameters for generating responses, except for GLM-4-9B, which was run on a machine with one Intel Xeon Platinum 8352V 2.10 GHz CPU and two NVIDIA GeForce RTX 4090 GPUs. For GPT-4o, we used the version \texttt{gpt-4o-2024-08-06}, and for Claude 3.5 Sonnet, we used the version \texttt{claude-3-5-sonnet-20240620}. Total cost for the experiments (including the key information extraction when curating the dataset) was approximately \$1500, with the majority of the cost attributed to the usage of GPT-4o, Claude 3.5 Sonnet, and GLM-4-Plus.

\section{Dataset Construction}
\subsection{Key Information Extraction Prompt}
\label{subsec:extraction_prompt}
We asked GPT-4o to extract the meaning, origin, and example sentences from the crawled meme explanation using the following prompt:
\begin{CJK*}{UTF8}{gbsn}
    \begin{tcolorbox}[breakable,enhanced]
        你需要根据提供的互联网流行梗的解释，提取它的含义、出处和\,3\,个例句。在提取时，保留所有关键信息，不要过度缩略。\textit{(You need to extract the meaning, origin, and three examples of usage based on the explanation of the provided Internet meme. When extracting, retain all key information without excessive abbreviation.)}
    \end{tcolorbox}
\end{CJK*}

\subsection{Manual Annotation}
\label{subsec:manual_annotation}
For the annotation of profanity, offense, and meme type, one of the authors conducted the initial annotation, which was then verified by the other two authors. When disagreements arose, the three authors held discussions to resolve any discrepancies, with final determinations made via majority voting. This process yielded perfect agreement (100\%) for profanity and offense classifications, and near-perfect agreement (92.9\%) for meme type. The final labels were agreed upon by all three authors, ensuring a consistent and accurate representation of the dataset.

During the pilot annotation phase of meme types, we initially followed a taxonomy derived from Chinese literature\footnote{\url{https://www.cpd.com.cn/n15737398/n26490099/523/t_1086228.html}} on Internet memes, which classified memes into four categories based on their creation methodology: \textit{phonetic}, \textit{experiential}, \textit{story-based}, and \textit{sarcastic}. However, when conducting comprehensive annotation of all memes, we discovered that this existing taxonomy could not adequately capture the full range of types we encountered. Consequently, we refined this classification system and expanded it to include two additional categories, resulting in a more comprehensive taxonomy that better represents the diversity of Chinese Internet memes.

\subsection{Statistics on Memes' Origins}
\label{subsec:origin_statistics}
Most memes with clear origins fall into the \textit{quotation} category (as expected). Of the total 1,458 memes, 525 have clear origins. We have compiled detailed statistics in Table~\ref{tab:origin_statistics}.
\begin{table}[htbp]
    \centering
    \begin{tabular}{lr}
        \toprule
        \textbf{Meme Type} & \textbf{\# Memes with Origin} \\
        \midrule
        Quotation & 411 (78.3\%) \\
        Homophonic pun & 41$\phantom{1}$ (7.8\%) \\
        Experience & 36$\phantom{1}$ (6.9\%) \\
        Stylistic device & 27$\phantom{1}$ (5.1\%) \\
        Abbreviation & 8$\phantom{1}$ (1.5\%) \\
        Slang & 2$\phantom{1}$ (0.4\%) \\
        \bottomrule
    \end{tabular}
    \caption{Statistics on the origins of memes in the CHIME dataset.}
    \label{tab:origin_statistics}
\end{table}

\subsection{Examples of the CHIME Dataset}
\label{subsec:chime_examples}
Table~\ref{tab:chime_examples_1} and Table~\ref{tab:chime_examples_2} provide a few representative examples of memes illustrating each meme type in the CHIME dataset.
\begin{table*}[htbp]
    \centering
    \includegraphics{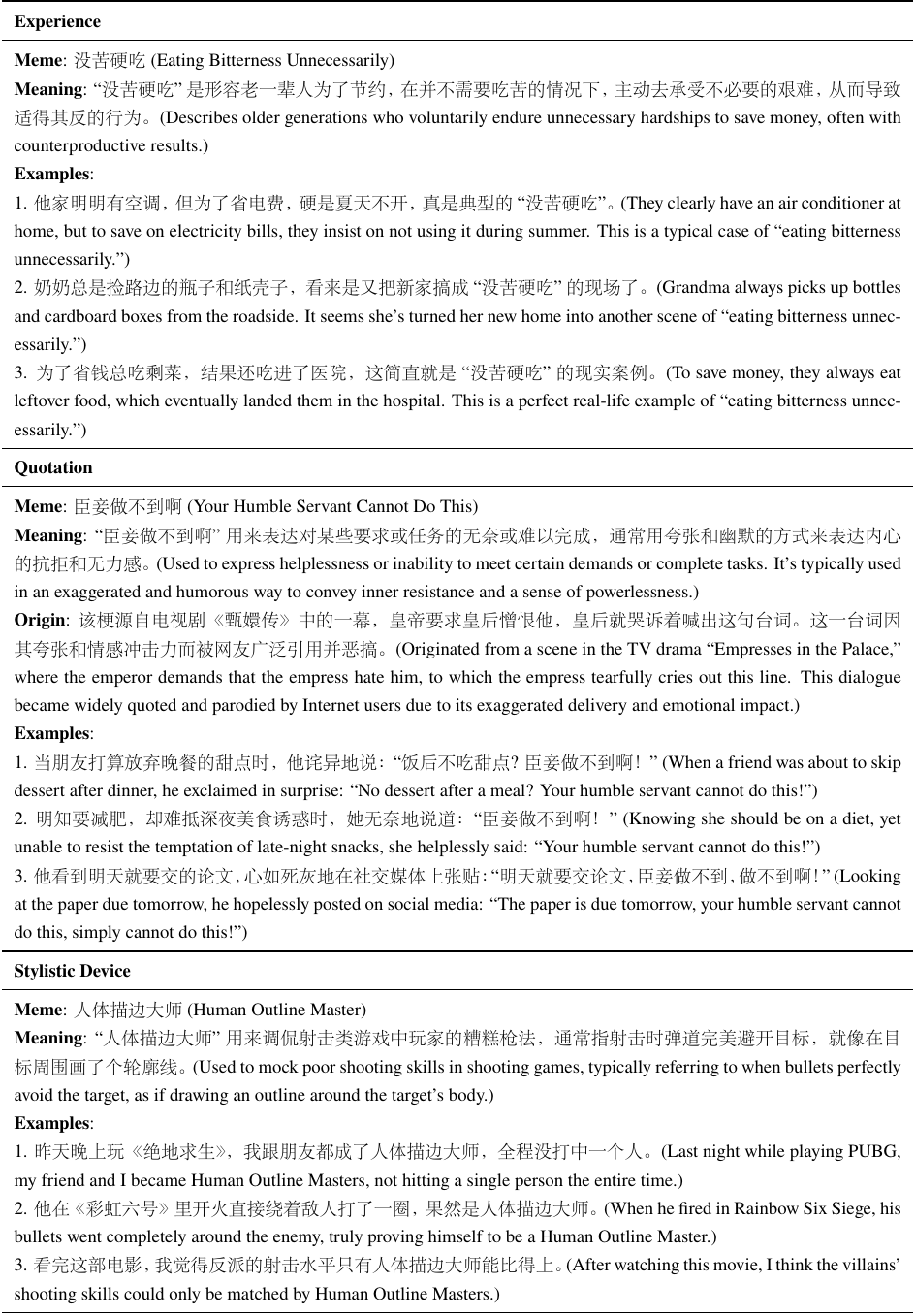}
    \caption{Representative examples from the CHIME dataset, illustrating the \textit{experience}, \textit{quotation}, and \textit{stylistic device} meme types.}
    \label{tab:chime_examples_1}
\end{table*}
\begin{table*}[htbp]
    \centering
    \includegraphics{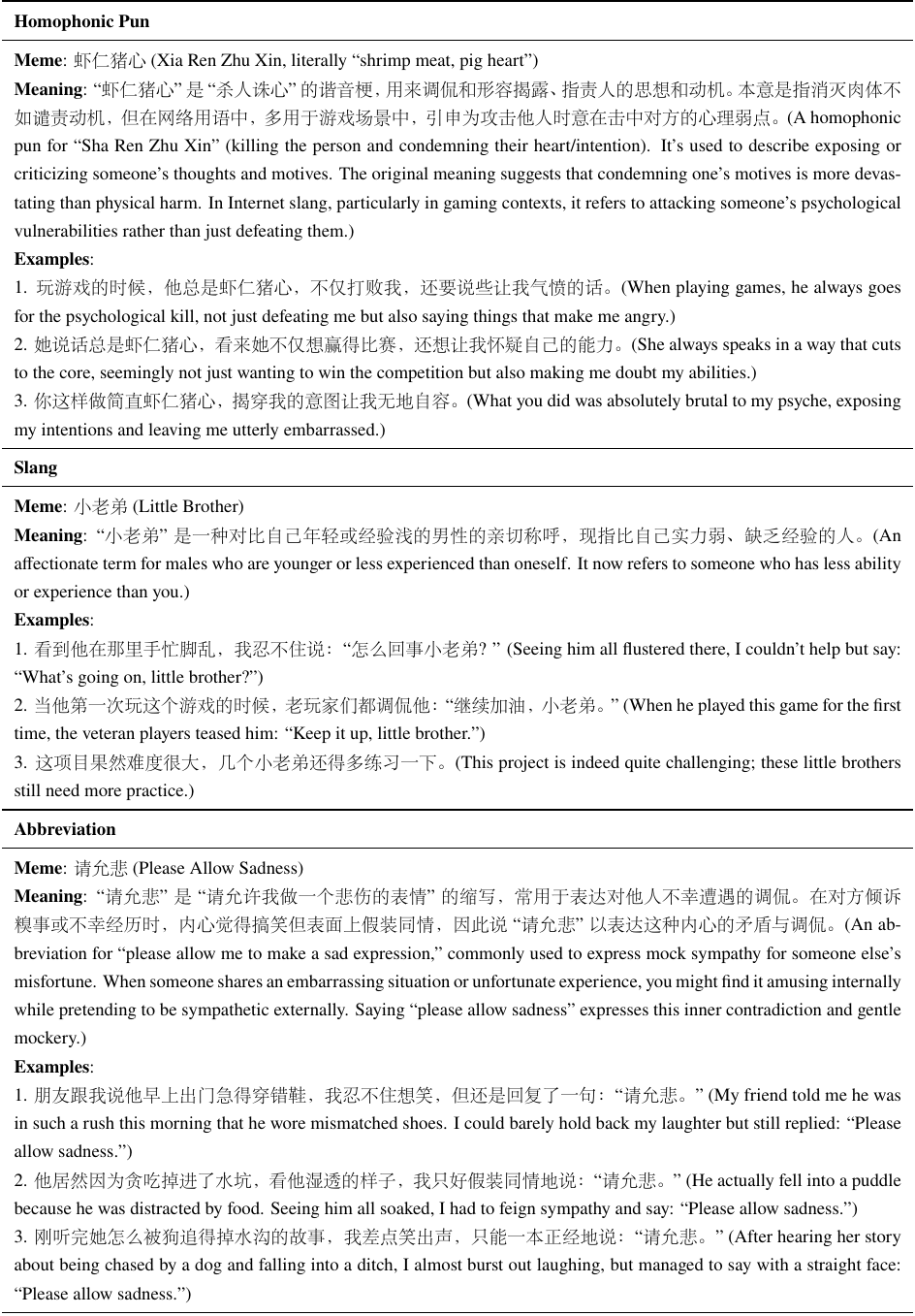}
    \caption{Representative examples from the CHIME dataset, illustrating the \textit{homophonic pun}, \textit{slang}, and \textit{abbreviation} meme types.}
    \label{tab:chime_examples_2}
\end{table*}

\section{Explanation Task}
\subsection{Zero-Shot Prompts}
\label{subsec:zero_shot_prompts}
We gave the following zero-shot prompts to the candidate models and let them explain the meaning of a given Internet meme, provide its origin (if available), and construct an example sentence:
\begin{CJK*}{UTF8}{gbsn}
    \begin{tcolorbox}[breakable,enhanced]
        \textit{For memes without a known origin:} \\[0.25em]
        在中文互联网的语境下，解释以下网络流行梗的含义，并撰写\,1\,个例句。\textit{(In the context of the Chinese Internet, explain the meaning of the following viral meme and create one example sentence.)} \\[0.5em]
        \textit{For memes with a known origin:} \\[0.25em]
        在中文互联网的语境下，解释以下网络流行梗的含义和出处，并撰写\,1\,个例句。\textit{(In the context of the Chinese Internet, explain the meaning and origin of the following viral meme, and create one example sentence.)}
    \end{tcolorbox}
\end{CJK*}

\subsection{One-Shot Prompts}
\label{subsec:one_shot_prompts}
We also experimented with one-shot prompts, where we provided the model with an example of a meme and its explanation, (possibly) origin, and example sentence:
\begin{CJK*}{UTF8}{gbsn}
    \begin{tcolorbox}[breakable,enhanced]
        \textit{For memes without a known origin:} \\[0.25em]
        在中文互联网的语境下，解释以下网络流行梗的含义，并撰写\,1\,个例句。\\
        示例：\\
        技术宅拯救世界\\
        含义：指技术宅能够通过自身强大的动手和创造能力，解决各种实际问题，甚至能承担起拯救世界的重任。\\
        例句：在电影里，当病毒席卷全球，终究是技术宅拯救世界，用代码解开谜团。\\
        \textit{(English translation)}\\
        \textit{In the context of the Chinese Internet, explain the meaning of the following viral meme and create one example sentence.}\\
        \textit{Example:}\\
        \textit{Tech Geeks Save the World}\\
        \textit{Meaning: Refers to how tech enthusiasts can solve various practical problems through their strong hands-on and creative abilities, and even take on the responsibility of saving the world.}\\
        \textit{Example sentence: In the movie, when the virus swept across the globe, it was ultimately the tech geeks who saved the world by decoding the mystery with their programming skills.}\\[0.5em]
        \textit{For memes with a known origin:} \\[0.25em]
        在中文互联网的语境下，解释以下网络流行梗的含义和出处，并撰写\,1\,个例句。\\
        示例：\\
        水灵灵\\
        含义：形容一种年轻、有活力的状态。\\
        出处：源自一位韩国女子组合成员在采访中的发言，她在展示合照封面时说自己\,“水灵灵地在中间”。\\
        例句：水灵灵地挤个地铁，每天都充满活力。\\
        \textit{(English translation)}\\
        \textit{In the context of the Chinese Internet, explain the meaning and origin of the following viral meme, and create one example sentence.}\\
        \textit{Example:}\\
        \textit{Fresh and Dewy}\\
        \textit{Meaning: Describes a youthful, energetic state or condition.}\\
        \textit{Origin: Originated from a statement made by a Korean girl group member during an interview, where she described herself as ``fresh and dewy in the middle'' when showing a group photo cover.}\\
        \textit{Example sentence: Taking the subway with a fresh and dewy attitude, filled with vitality every day.}
    \end{tcolorbox}
\end{CJK*}

Our analysis revealed that one-shot prompts did not significantly improve model performance, but greatly diminished it on the meaning explanation task, compared to zero-shot prompts, as demonstrated in Table~\ref{tab:zero_shot_vs_one_shot}. We hypothesize that this performance degradation stems from the inherent nature of meme interpretation, which demands flexible analysis rather than rigid pattern matching or format adherence. Consequently, we focused exclusively on zero-shot prompting results in the main text.
\begin{table*}[htbp]
    \centering
    \begin{tabular}{lccccccccc}
        \toprule
        && \multicolumn{2}{c}{\textbf{Cosine Similarity}} && \multicolumn{2}{c}{\textbf{BERTScore (F)}} && \multicolumn{2}{c}{\textbf{BARTScore (F)}} \\
        \cmidrule{3-4} \cmidrule{6-7} \cmidrule{9-10}
        \textbf{Model} && \textbf{Meaning} & \textbf{Origin} && \textbf{Meaning} & \textbf{Origin} && \textbf{Meaning} & \textbf{Origin} \\
        \midrule
        GPT-4o && & && & && & \\
        \quad Zero-Shot && 0.815 $\phantom{\uparrow}$ & 0.647 $\phantom{\uparrow}$ && 0.800 $\phantom{\uparrow}$ & 0.675 $\phantom{\uparrow}$ && $-$4.485 $\phantom{\uparrow}$ & $-$4.717 $\phantom{\uparrow}$ \\
        \quad One-Shot && 0.825 $\uparrow$ & 0.652 $\uparrow$ && 0.805 $\uparrow$ & 0.717 $\uparrow$ && $-$4.426 $\uparrow$ & $-$4.565 $\uparrow$ \\
        \midrule
        Claude 3.5 Sonnet && & && & && & \\
        \quad Zero-Shot && 0.788 $\phantom{\uparrow}$ & 0.625 $\phantom{\uparrow}$ && 0.789 $\phantom{\uparrow}$ & 0.696 $\phantom{\uparrow}$ && $-$4.611 $\phantom{\uparrow}$ & $-$4.695 $\phantom{\uparrow}$ \\
        \quad One-Shot && 0.736 $\downarrow$ & 0.660 $\uparrow$ && 0.761 $\downarrow$ & 0.719 $\uparrow$ && $-$4.630 $\downarrow$ & $-$4.750 $\downarrow$ \\
        \midrule
        GLM-4-9B && & && & && & \\
        \quad Zero-Shot && 0.813 $\phantom{\uparrow}$ & 0.578 $\phantom{\uparrow}$ && 0.797 $\phantom{\uparrow}$ & 0.663 $\phantom{\uparrow}$ && $-$4.453 $\phantom{\uparrow}$ & $-$4.560 $\phantom{\uparrow}$ \\
        \quad One-Shot && 0.750 $\downarrow$ & 0.549 $\downarrow$ && 0.746 $\downarrow$ & 0.607 $\downarrow$ && $-$4.470 $\downarrow$ & $-$4.656 $\downarrow$ \\
        \midrule
        GLM-4-Plus && & && & && & \\
        \quad Zero-Shot && 0.844 $\phantom{\uparrow}$ & 0.679 $\phantom{\uparrow}$ && 0.822 $\phantom{\uparrow}$ & 0.737 $\phantom{\uparrow}$ && $-$4.291 $\phantom{\uparrow}$ & $-$4.441 $\phantom{\uparrow}$ \\
        \quad One-Shot && 0.797 $\downarrow$ & 0.689 $\uparrow$ && 0.796 $\downarrow$ & 0.743 $\uparrow$ && $-$4.283 $\uparrow$ & $-$4.468 $\downarrow$ \\
        \midrule
        Qwen2.5-7B && & && & && & \\
        \quad Zero-Shot && 0.792 $\phantom{\uparrow}$ & 0.605 $\phantom{\uparrow}$ && 0.782 $\phantom{\uparrow}$ & 0.661 $\phantom{\uparrow}$ && $-$4.494 $\phantom{\uparrow}$ & $-$4.779 $\phantom{\uparrow}$ \\
        \quad One-Shot && 0.731 $\downarrow$ & 0.639 $\uparrow$ && 0.731 $\downarrow$ & 0.693 $\uparrow$ && $-$4.573 $\downarrow$ & $-$4.677 $\uparrow$ \\
        \midrule
        Qwen2.5-72B && & && & && & \\
        \quad Zero-Shot && 0.819 $\phantom{\uparrow}$ & 0.627 $\phantom{\uparrow}$ && 0.803 $\phantom{\uparrow}$ & 0.690 $\phantom{\uparrow}$ && $-$4.366 $\phantom{\uparrow}$ & $-$4.605 $\phantom{\uparrow}$ \\
        \quad One-Shot && 0.799 $\downarrow$ & 0.626 $\downarrow$ && 0.789 $\downarrow$ & 0.697 $\uparrow$ && $-$4.370 $\downarrow$ & $-$4.498 $\uparrow$ \\
        \midrule
        DeepSeek-V3 && & && & && & \\
        \quad Zero-Shot && 0.779 $\phantom{\uparrow}$ & 0.709 $\phantom{\uparrow}$ && 0.774 $\phantom{\uparrow}$ & 0.751 $\phantom{\uparrow}$ && $-$4.331 $\phantom{\uparrow}$ & $-$4.344 $\phantom{\uparrow}$ \\
        \quad One-Shot && 0.746 $\downarrow$ & 0.689 $\downarrow$ && 0.754 $\downarrow$ & 0.722 $\downarrow$ && $-$4.380 $\downarrow$ & $-$4.539 $\downarrow$ \\
        \bottomrule
    \end{tabular}
    \caption{Comparative analysis of average cosine similarity, BERTScore, and BARTScore across six meme types for all candidate models, contrasting zero-shot and one-shot prompting approaches. $\uparrow$ indicates superior performance, and $\downarrow$ denotes inferior performance. Results were derived from a balanced sample of 240 memes, comprising 40 from each meme type.}
    \label{tab:zero_shot_vs_one_shot}
\end{table*}

\subsection{More Automatic Evaluation Results}
\label{subsec:auto_eval_results}
\begin{table*}[htbp]
    \centering
    \begin{tabular}{lcccccccc}
        \toprule
        && \multicolumn{3}{c}{\textbf{Experience}} && \multicolumn{3}{c}{\textbf{Quotation}} \\
        \cmidrule{3-5} \cmidrule{7-9}
        \textbf{Model} && \textbf{Cos.~Sim.} & \textbf{BERTS.} & \textbf{BARTS.} && \textbf{Cos.~Sim.} & \textbf{BERTS.} & \textbf{BARTS.} \\
        \midrule
        GPT-4o && 0.837 & 0.812 & $-$4.261 && 0.756 & 0.755 & $-$4.354 \\
        Claude 3.5 Sonnet && 0.809 & 0.799 & $-$4.414 && 0.707 & 0.733 & $-$4.594 \\
        GLM-4-9B && 0.818 & 0.804 & $-$4.236 && 0.740 & 0.750 & $-$4.285 \\
        GLM-4-Plus && \cyanbox{\textbf{0.846}} & \cyanbox{\textbf{0.822}} & \cyanbox{$\bm{-}$\textbf{4.150}} && \cyanbox{\textbf{0.812}} & \cyanbox{\textbf{0.790}} & \cyanbox{$\bm{-}$\textbf{4.199}} \\
        Qwen2.5-7B && 0.807 & 0.786 & $-$4.337 && 0.731 & 0.730 & $-$4.447 \\
        Qwen2.5-72B && 0.832 & 0.807 & $-$4.220 && 0.763 & 0.757 & $-$4.294 \\
        DeepSeek-V3 && 0.802 & 0.796 & $-$4.203 && 0.742 & 0.742 & $-$4.306 \\

        \bottomrule \\[-1em]

        && \multicolumn{3}{c}{\textbf{Stylistic Device}} && \multicolumn{3}{c}{\textbf{Homophonic Pun}} \\
        \cmidrule{3-5} \cmidrule{7-9}
        \textbf{Model} && \textbf{Cos.~Sim.} & \textbf{BERTS.} & \textbf{BARTS.} && \textbf{Cos.~Sim.} & \textbf{BERTS.} & \textbf{BARTS.} \\
        \midrule
        GPT-4o && 0.811 & 0.792 & $-$4.365 && 0.797 & 0.789 & $-$4.751 \\
        Claude 3.5 Sonnet && 0.790 & 0.782 & $-$4.499 && 0.781 & 0.784 & $-$5.101 \\
        GLM-4-9B && 0.805 & 0.791 & $-$4.303 && 0.799 & 0.790 & $-$4.741 \\
        GLM-4-Plus && \cyanbox{\textbf{0.827}} & \cyanbox{\textbf{0.804}} & $-$4.248 && \cyanbox{\textbf{0.826}} & \cyanbox{\textbf{0.808}} & \cyanbox{$\bm{-}$\textbf{4.589}} \\
        Qwen2.5-7B && 0.791 & 0.771 & $-$4.387 && 0.762 & 0.761 & $-$4.875 \\
        Qwen2.5-72B && 0.813 & 0.793 & $-$4.312 && 0.794 & 0.788 & $-$4.746 \\
        DeepSeek-V3 && 0.802 & 0.794 & \cyanbox{$\bm{-}$\textbf{4.245}} && 0.818 & 0.805 & $-$4.610 \\

        \bottomrule \\[-1em]

        && \multicolumn{3}{c}{\textbf{Slang}} && \multicolumn{3}{c}{\textbf{Abbreviation}} \\
        \cmidrule{3-5} \cmidrule{7-9}
        \textbf{Model} && \textbf{Cos.~Sim.} & \textbf{BERTS.} & \textbf{BARTS.} && \textbf{Cos.~Sim.} & \textbf{BERTS.} & \textbf{BARTS.} \\
        \midrule
        GPT-4o && 0.835 & 0.809 & $-$4.388 && 0.830 & 0.819 & $-$4.612 \\
        Claude 3.5 Sonnet && 0.802 & 0.791 & $-$4.531 && 0.820 & 0.815 & $-$4.736 \\
        GLM-4-9B && 0.827 & 0.807 & \cyanbox{$\bm{-}$\textbf{4.253}} && 0.845 & 0.826 & $-$4.607 \\
        GLM-4-Plus && 0.837 & 0.812 & $-$4.332 && \cyanbox{\textbf{0.865}} & \cyanbox{\textbf{0.839}} & $-$4.479 \\
        Qwen2.5-7B && 0.814 & 0.792 & $-$4.415 && 0.792 & 0.786 & $-$4.863 \\
        Qwen2.5-72B && \cyanbox{\textbf{0.837}} & \cyanbox{\textbf{0.818}} & $-$4.290 && 0.830 & 0.816 & $-$4.636 \\
        DeepSeek-V3 && 0.800 & 0.798 & $-$4.272 && 0.840 & 0.824 & \cyanbox{$\bm{-}$\textbf{4.467}} \\
        \bottomrule
    \end{tabular}
    \caption{Average cosine similarity, BERTScore, and BARTScore for the generated meanings of the candidate models, for each of the six meme types. The best-performing scores are highlighted in \cyanbox{\textbf{bold}}.}
    \label{tab:auto_eval_results_more}
\end{table*}
Table~\ref{tab:auto_eval_results_more} gives the exact meaning scores of the candidate models for each of the six meme types.

\subsection{Human Evaluation Details}
\label{subsec:human_eval_instruct_example}
For our human evaluation process, we first divided the 240 testing memes into 12 batches of 20 memes each. For each batch, we created a questionnaire containing an instruction page followed by 20 evaluation pages (one per meme). The instruction page provided the following guidelines to raters (translated from Chinese):
\begin{tcolorbox}[breakable,enhanced]
    Internet memes, as a unique cultural phenomenon, not only reflect societal trends and public emotions but also hold significant social influence. To study the understanding of Chinese Internet memes by large language models, this project aims to systematically evaluate Internet memes within the context of the Chinese Internet through a questionnaire survey.\\[0.5\baselineskip]
    This questionnaire is divided into two parts: The first part will collect your name; the second part consists of 20 pages, each corresponding to one popular meme. You will be required to evaluate the explanations of each meme generated by six large language models across three dimensions: ``meaning,'' ``origin,'' and ``example sentence.''\\[0.5\baselineskip]
    You will answer approximately 120 questions, and the survey is expected to take about 40 minutes.\\[0.5\baselineskip]
    \textbf{I. Instructions}
    \begin{enumerate}
        \item Participation in this survey is entirely voluntary. You have the right to decide whether to participate. Your personal information will be kept strictly confidential and used solely for academic research purposes, with no disclosure to third parties.
        \item To ensure the accuracy and reliability of the survey results, please provide honest answers and avoid random responses or providing false information.
        \item Please complete the questionnaire to the fullest extent possible and avoid skipping any questions. If you have any doubts, feel free to contact the project team for clarification.
        \item Once you have completed the questionnaire, click the ``Submit'' button to confirm your submission. Please note that submissions cannot be modified, so review your responses carefully before submitting.
        \item Be advised that the questionnaire may contain some vulgar, sexually suggestive, or offensive content. If you feel uncomfortable with such content, please consider whether to proceed.
    \end{enumerate}
    \textbf{II. Acknowledgments and Feedback}
    \begin{enumerate}
        \item Thank you for taking the time to participate in this survey. Every response you provide will contribute valuable data to our research.
        \item If you encounter any issues or have any suggestions while filling out the questionnaire, feel free to contact the project team at any time.
        \item After the survey is complete, the project team will analyze the data and prepare a research report. If needed, we will share the results of the study with participants.
    \end{enumerate}
    Thank you once again for your support and cooperation!
\end{tcolorbox}

For each questionnaire, ratings were collected from three independent raters. We payed each rater around \$14 per hour for their participation, which is much higher than the average hourly wage in China. We reruited a total number of 14 raters for the human evaluation task, and their birth years range from 1980s to 2000s. All raters were native Chinese speakers with a good understanding of Chinese Internet culture. Of the 14 raters, 9 annotated three batches, 4 annotated two batches, and 1 annotated a single batch. The average number of batches per rater was 2.57, with a median of 3.

\subsection{More Human Evaluation Results}
\label{subsec:human_eval_results}
Table~\ref{tab:fleiss_kappa} gives the Fleiss' kappa scores on each of the 12 evaluation batches. Based on the kappa scores, we observe that \emph{abbreviation} memes show highest agreement ($\kappa\approx 0.71$ to $0.74$) due to their straightforward nature; \emph{slang} memes show lowest agreement ($\kappa\approx 0.27$ to $0.28$) because cultural familiarity varies among annotators; \emph{homophonic puns} show moderate disagreement ($\kappa\approx 0.40$ to $0.41$) due to subjective interpretation of wordplay effectiveness. We conjecture that cultural context dependency is the primary driver of annotation disagreement---memes requiring deeper cultural knowledge (slang, stylistic devices) are harder to evaluate consistently than structurally-defined ones (abbreviations). Table~\ref{tab:human_eval_results_more} provides the detailed human evaluation results on the meaning task for each of the six meme types.
\begin{table}[htbp]
    \centering
    \begin{tabular}{clc}
        \toprule
        \textbf{Batch} & \textbf{Meme Type} & \textbf{Fleiss' kappa} \\
        \midrule
        1 & Slang & 0.278 \\
        2 & Slang & 0.269 \\
        3 & Stylistic device & 0.318 \\
        4 & Stylistic device & 0.487 \\
        5 & Quotation & 0.421 \\
        6 & Quotation & 0.519 \\
        7 & Experience & 0.360 \\
        8 & Experience & 0.393 \\
        9 & Abbreviation & 0.736 \\
        10 & Abbreviation & 0.711 \\
        11 & Homophonic pun & 0.412 \\
        12 & Homophonic pun & 0.400 \\
        \bottomrule
    \end{tabular}
    \caption{Fleiss' kappa scores on each of the 12 evaluation batches in human evaluation.}
    \label{tab:fleiss_kappa}
\end{table}
\begin{table*}[htbp]
    \centering
    \begin{tabular}{lcccccccccccc}
        \toprule
        && \multicolumn{3}{c}{\textbf{Experience (\%)}} && \multicolumn{3}{c}{\textbf{Quotation (\%)}} && \multicolumn{3}{c}{\textbf{Stylistic Device (\%)}} \\
        \cmidrule{3-5} \cmidrule{7-9} \cmidrule{11-13}
        \textbf{Model} && \textbf{A} & \textbf{N} & \textbf{D} && \textbf{A} & \textbf{N} & \textbf{D} && \textbf{A} & \textbf{N} & \textbf{D} \\
        \midrule
        GPT-4o && 70.8 & 5.9 & 23.3 && 35.8 & 10.9 & 53.3 && 65.0 & 7.5 & 27.5 \\
        Claude 3.5 Sonnet && 67.5 & 6.7 & 25.8 && 34.2 & 8.3 & 57.5 && 50.8 & 12.5 & 36.7 \\
        GLM-4-9B && 61.6 & 1.7 & 36.7 && 20.8 & 15.9 & 63.3 && 42.5 & 8.3 & 49.2 \\
        GLM-4-Plus && \tealbox{\textbf{80.8}} & 3.4 & 15.9 && 48.3 & 15.8 & 35.8 && 69.1 & 9.2 & \tealbox{\textbf{21.7}} \\
        Qwen2.5-7B && 47.5 & 14.2 & 38.3 && 20.8 & 6.7 & 72.5 && 32.5 & 12.5 & 55.0 \\
        Qwen2.5-72B && 64.2 & 3.3 & 32.5 && 22.5 & 15.8 & 61.7 && 50.8 & 12.5 & 36.7 \\
        DeepSeek-V3 && 77.5 & 15.0 & \tealbox{\textbf{7.5}} && \tealbox{\textbf{70.8}} & 11.7 & \tealbox{\textbf{17.5}} && \tealbox{\textbf{73.3}} & 3.4 & 23.3 \\

        \bottomrule \\[-1em]

        && \multicolumn{3}{c}{\textbf{Homophonic Pun (\%)}} && \multicolumn{3}{c}{\textbf{Slang (\%)}} && \multicolumn{3}{c}{\textbf{Abbreviation (\%)}} \\
        \cmidrule{3-5} \cmidrule{7-9} \cmidrule{11-13}
        \textbf{Model} && \textbf{A} & \textbf{N} & \textbf{D} && \textbf{A} & \textbf{N} & \textbf{D} && \textbf{A} & \textbf{N} & \textbf{D} \\
        \midrule
        GPT-4o && 32.5 & 11.7 & 55.8 && 77.5 & 10.8 & 11.7 && 41.7 & 7.5 & 50.8 \\
        Claude 3.5 Sonnet && 29.2 & 14.2 & 56.6 && 79.1 & 9.2 & 11.7 && 45.0 & 7.5 & 47.5 \\
        GLM-4-9B && 12.5 & 12.5 & 75.0 && 75.0 & 10.0 & 15.0 && 30.0 & 5.8 & 64.2 \\
        GLM-4-Plus && 59.2 & 13.3 & \tealbox{\textbf{27.5}} && 85.8 & 8.4 & 5.8 && 67.5 & 3.3 & 29.2 \\
        Qwen2.5-7B && 19.2 & 10.8 & 70.0 && 60.8 & 15.0 & 24.2 && 22.5 & 9.2 & 68.3 \\
        Qwen2.5-72B && 20.8 & 15.9 & 63.3 && 76.6 & 11.7 & 11.7 && 39.2 & 0.8 & 60.0 \\
        DeepSeek-V3 && \tealbox{\textbf{60.0}} & 10.8 & 29.2 && \tealbox{\textbf{88.3}} & 9.2 & \tealbox{\textbf{2.5}} && \tealbox{\textbf{71.6}} & 11.7 & \tealbox{\textbf{16.7}} \\
        \bottomrule
    \end{tabular}
    \caption{Average percentage of human ratings assigned as \textit{Agree}, \textit{Neutral}, and \textit{Disagree} of the candidate models for each meme type, on the meaning task. A stands for \textit{Agree}, N stands for \textit{Neutral}, and D stands for \textit{Disagree}. The best-performing scores are highlighted in \tealbox{\textbf{bold}}.}
    \label{tab:human_eval_results_more}
\end{table*}

\subsection{Error Analysis with Illustrative Case Studies}
\label{subsec:error_analysis}
To further investigate the performance of LLMs, we conducted a qualitative error analysis on the generated meanings and origins. Specifically, we have identified three common types of errors in the generated meanings and origins, which are as follows:
\begin{CJK*}{UTF8}{gbsn}
\begin{enumerate}
    \item \textbf{Origin confusion}: Models frequently attributed memes to incorrect sources, particularly with quotation memes. In many instances, LLMs provided vague attributions (e.g., ``originating from social media'') rather than specific origins. For example, 再爱就不礼貌了\ (Any More Love Would Be Impolite) originated when a Japanese short video blogger made it into a Japanese language teaching video, which Internet users then parodied into a ``fake Japanese version,'' creating a comedic atmosphere. However, all models except Qwen2.5-7B and DeepSeek-V3 have provided vague origins, such as ``the exact origin of this meme is unclear, but it gradually gained popularity in social media and everyday online communication'' (by GLM-4-Plus), while Qwen2.5-7B and DeepSeek-V3 provided completely incorrect origins.
    \item \textbf{Semantic shift}: For most misinterpreted homophonic pun memes, models explained related concepts with similar phonetics rather than capturing the actual meme meaning. In other cases, models failed to recognize the phonetic wordplay entirely and simply explained the literal meaning. For example, 肾炎\ (Shen Yan, literally ``nephritis'') is a homophonic pun on ``divine face/godly appearance'' in Chinese, used to mock fans who exaggeratedly describe their idols as having ``godly looks.'' However, Claude 3.5 Sonnet misinterpreted it as another homophonic word 神言, which means ``divine words'' (words of god/godlike statement), while GLM-4-Plus simply explained its literal meaning, interpreting it as a kidney disease.
    \item \textbf{Cross-type confusion}: Abbreviation memes were occasionally misinterpreted as homophonic puns, indicating difficulty in distinguishing between these distinct linguistic mechanisms. For example, 人干事\ (Human Doing Things) is an abbreviation of 这是人干的事吗\ (Is This Something A Human Would Do?), mainly used to criticize unreasonable or unacceptable things. However, Qwen2.5-7B recognized it as a homophonic pun, 人设\ (Persona), and explained ``used to mock people whose online image is inconsistent with their actual behavior.''
\end{enumerate}
\end{CJK*}
Table~\ref{tab:error_analysis_outputs} gives the complete model outputs for the above three error types.
\begin{table*}[htbp]
    \centering
    \includegraphics{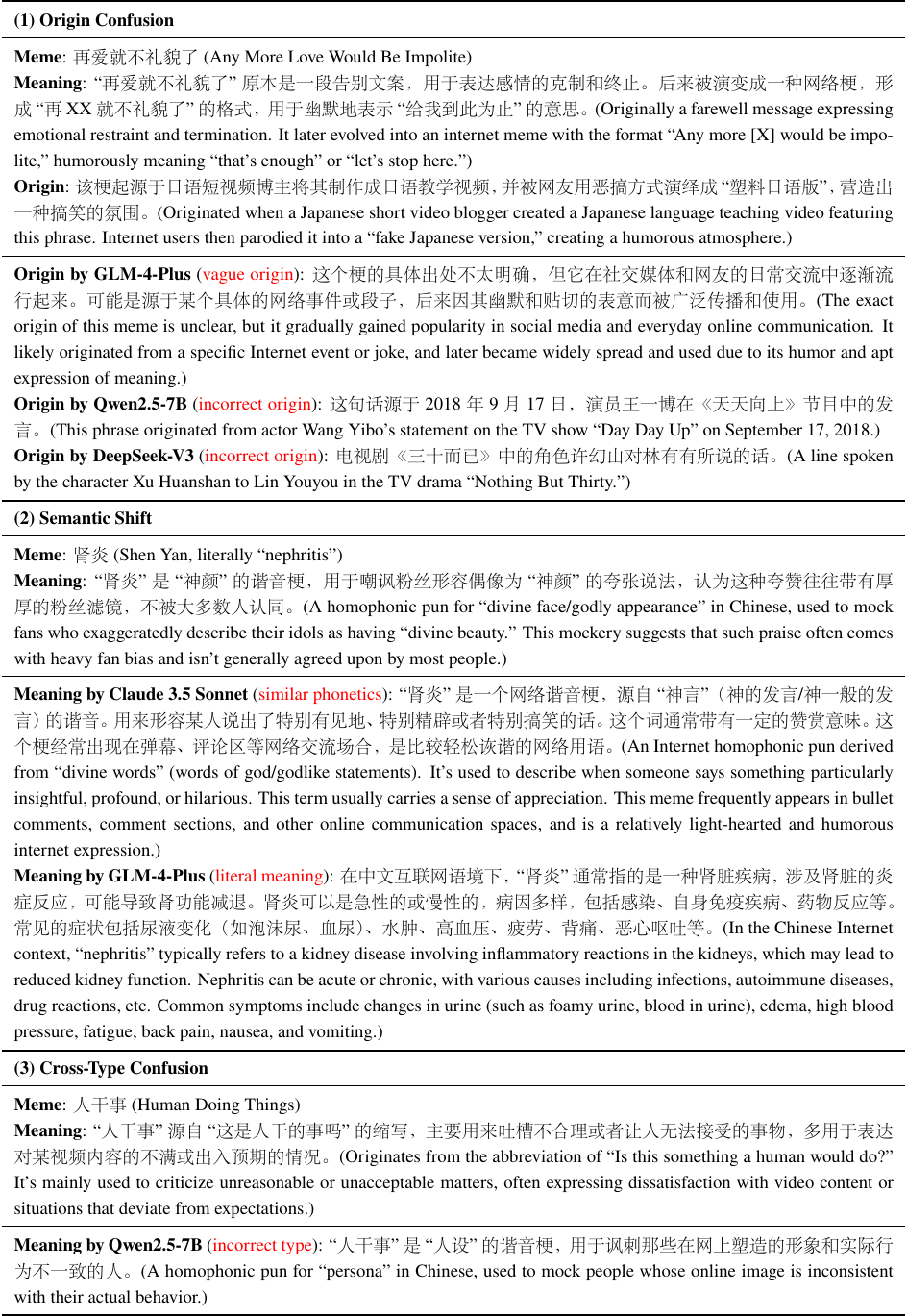}
    \caption{Examples of model outputs for three common error types: (1) Origin confusion, (2) Semantic shift, and (3) Cross-type confusion.}
    \label{tab:error_analysis_outputs}
\end{table*}

\section{MCQ Task}
\subsection{MCQ Prompt without Meaning}
\label{subsec:mcq_prompts_without_meaning}
For the multiple-choice questions (MCQs), we provided the following prompts to the candidate models (with English translation):
\begin{CJK*}{UTF8}{gbsn}
    \begin{tcolorbox}[breakable,enhanced]
        根据提供的句子，其中包含一个空白处，请从提供的\,5\,个选项中，根据上下文选择最合适的网络流行梗填入。只需给出选项的编号作为答案，不要做任何解释。\\
        示例：\\
        句子：这个方案真是\_\_\_\_\_，完全超出我的想象。\\
        选项：\\
        (1) 雪糕刺客\\
        (2) yyds\\
        (3) 狗带\\
        (4) 实锤\\
        (5) 偷感很重\\
        答案：2\\[0.5em]
        \itshape
        (English translation)\\
        Based on the given sentence, which contains a blank, choose the most suitable Internet meme from the five provided options according to the context. Only provide the option number as the answer, without any explanation.\\
        Example:\\
        Sentence: This plan is truly \_\_\_\_\_, completely beyond my imagination.\\
        Options:\\
        (1) Ice Cream Assassin\\
        (2) yyds (similar to GOAT in English)\\
        (3) Go Die\\
        (4) Solid Evidence\\
        (5) Strong Sense of Stealing\\
        Answer: 2
    \end{tcolorbox}
\end{CJK*}

\subsection{MCQ Prompt with Meaning}
\label{subsec:mcq_prompts_with_meaning}
For MCQs where the meaning of each meme option was provided to the LLMs, the prompt was as follows (with English translation):
\begin{CJK*}{UTF8}{gbsn}
    \begin{tcolorbox}[breakable,enhanced]
        根据提供的句子，其中包含一个空白处，请从提供的\,5\,个选项中，根据上下文选择最合适的网络流行梗填入。只需给出选项的编号作为答案，不要做任何解释。\\
        示例：\\
        句子：这个方案真是\_\_\_\_\_，完全超出我的想象。\\
        选项：\\
        (1) 雪糕刺客。含义：“雪糕刺客”\,指的是那些看似普通但价格高昂的雪糕，购买时让人感到意外和“被刺”的疼痛感。这个表达反映了雪糕价格上涨和意外负担感。\\
        (2) yyds。含义：yyds\,是\,“永远的神”\,的缩写，用来称赞某人或某事物非常优秀，值得敬仰和追随。\\
        (3) 狗带。含义：“狗带”\,是\,“go die”\,的谐音，意为去死或者死亡，通常用于幽默或夸张的表达方式。\\
        (4) 实锤。含义：“实锤”\,指的是能够证明某事件真实发生的可靠证据，通常具备较强的说服力。\\
        (5) 偷感很重。含义：形容人在某些情境下感到拘谨、畏缩，显得偷偷摸摸或不自然。\\
        答案：2\\[0.5em]
        \textit{(English translation)}\\
        \textit{Based on the given sentence, which contains a blank, choose the most suitable Internet meme from the five provided options according to the context. Only provide the option number as the answer, without any explanation.}\\
        \textit{Example:}\\
        \textit{Sentence: This plan is truly \_\_\_\_\_, completely beyond my imagination.}\\
        \textit{Options:}\\
        \textit{(1) Ice Cream Assassin. Meaning: ``Ice Cream Assassin'' refers to seemingly ordinary but unexpectedly expensive ice cream, making people feel ``stabbed'' by the price. This phrase reflects rising ice cream prices and the unexpected financial burden.}\\
        \textit{(2) yyds. Meaning: ``yyds'' is the abbreviation for ``}永远的神\textit{'' (Eternal God), used to praise someone or something as excellent, admirable, and worthy of following.}\\
        \textit{(3) Go Die. Meaning: ``Go Die'' is a phonetic translation of ``}狗带\textit{'' (gǒu dài), meaning ``to die'' or ``go to hell,'' often used humorously or exaggeratedly.}\\
        \textit{(4) Solid Evidence. Meaning: ``Solid Evidence'' refers to strong and reliable proof that confirms an event or claim, typically carrying strong credibility.}\\
        \textit{(5) Strong Sense of Stealing. Meaning: This phrase describes someone feeling awkward, timid, or unnatural in a certain situation, appearing sneaky or out of place.}\\
        Answer: 2
    \end{tcolorbox}
\end{CJK*}

\end{document}